\documentclass[letterpaper, 10 pt, conference]{ieeeconf}  

\IEEEoverridecommandlockouts                              

\usepackage{multicol}
\usepackage{multirow}
\usepackage{float} 
\usepackage{siunitx} 
\usepackage{tablefootnote}
\usepackage{booktabs} 

\usepackage{svg}
\usepackage{graphicx}
\usepackage{subcaption}

\usepackage{amsmath}
\usepackage{gensymb}
\usepackage{amssymb} 
\usepackage{relsize} 

\usepackage{cite} 

\makeatletter
\let\NAT@parse\undefined
\makeatother
\usepackage[hidelinks]{hyperref}

\usepackage[capitalize]{cleveref}
\crefname{figure}{Fig.}{Figs.}
\Crefname{figure}{Figure}{Figures}
\crefname{table}{Table}{Tables}
\Crefname{table}{Table}{Tables}

\title{\LARGE \bf
HARP-NeXt: High-Speed and Accurate Range-Point Fusion Network for 3D LiDAR Semantic Segmentation}

\author{Samir Abou Haidar$^{1,2}$, Alexandre Chariot$^{1}$, Mehdi Darouich$^{1}$, Cyril Joly$^{2}$ and Jean-Emmanuel Deschaud$^{2}$
\thanks{$^{1}$The authors are with Paris-Saclay University, CEA, List, F-91120, Palaiseau, France
        {\tt\small first\_name.last\_name@cea.fr}}%
\thanks{$^{2}$The authors are with Mines Paris, PSL University, Centre for Robotics (CAOR), 75006 Paris, France
        {\tt\small first\_name.last\_name@minesparis.psl.eu}}%
}

\begin{document}

\maketitle
\thispagestyle{empty}
\pagestyle{empty}

\begin{abstract}
LiDAR semantic segmentation is crucial for autonomous vehicles and mobile robots, requiring high accuracy and real-time processing, especially on resource-constrained embedded systems. Previous state-of-the-art methods often face a trade-off between accuracy and speed. Point-based and sparse convolution-based methods are accurate but slow due to the complexity of neighbor searching and 3D convolutions. Projection-based methods are faster but lose critical geometric information during the 2D projection. Additionally, many recent methods rely on test-time augmentation (TTA) to improve performance, which further slows the inference. Moreover, the pre-processing phase across all methods increases execution time and is demanding on embedded platforms. Therefore, we introduce HARP-NeXt, a high-speed and accurate LiDAR semantic segmentation network. We first propose a novel pre-processing methodology that significantly reduces computational overhead. Then, we design the Conv-SE-NeXt feature extraction block to efficiently capture representations without deep layer stacking per network stage. We also employ a multi-scale range-point fusion backbone that leverages information at multiple abstraction levels to preserve essential geometric details, thereby enhancing accuracy. Experiments on the nuScenes and SemanticKITTI benchmarks show that HARP-NeXt achieves a superior speed-accuracy trade-off compared to all state-of-the-art methods, and, without relying on ensemble models or TTA, is comparable to the top-ranked PTv3, while running 24$\times$ faster. The code is available at \href{https://github.com/SamirAbouHaidar/HARP-NeXt}{https://github.com/SamirAbouHaidar/HARP-NeXt}
\end{abstract}
    
\section{INTRODUCTION}
\label{sec:introduction}
  

  
Autonomous vehicles and robots widely rely on LiDAR sensors to generate 3D point cloud data, enabling them to perceive and understand their surroundings with high precision. A key component for such systems is LiDAR semantic segmentation, which involves classifying each point in the raw point cloud into meaningful classes such as cars, pedestrians, roads, and buildings. This is fundamental for object detection and recognition, which are essential for safe autonomous navigation. In recent years, numerous deep learning models have been proposed to segment LiDAR 3D point clouds. However, these methods struggle to ensure both high accuracy and fast processing simultaneously, particularly on mobile autonomous and robotic systems that rely on resource-constrained computational platforms, such as NVIDIA Jetson devices. 

Although 3D point sets are rich in information, they inherently present several challenges, including an unstructured format, sparsity (spatial gaps), and substantial size. Methods
\begin{figure}[H]
  \centering
    \includegraphics[width=0.475\textwidth]{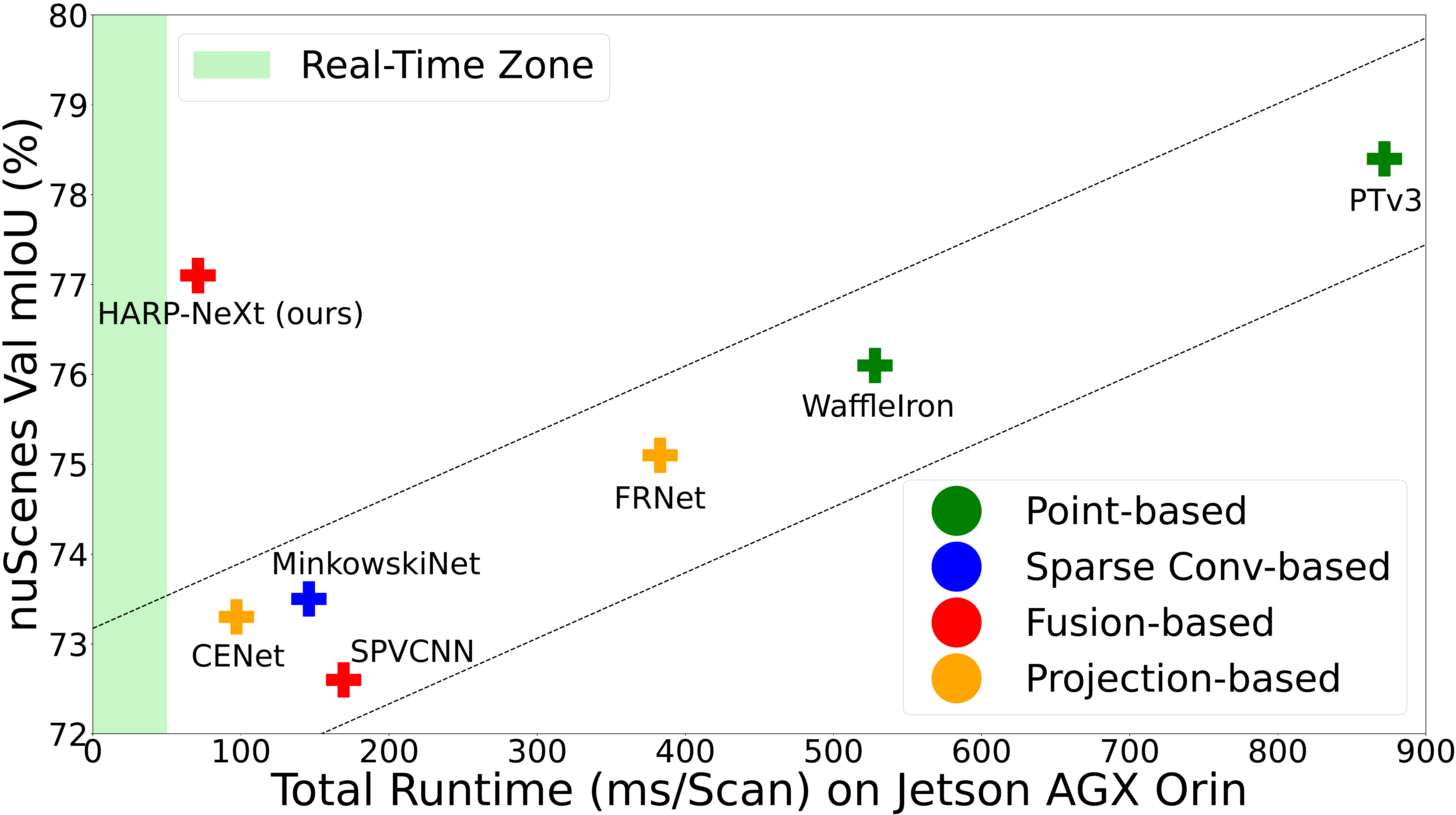}
    \caption{mIoU vs. runtime on nuScenes validation set. HARP-NeXt achieves high accuracy in near real-time with fast execution, breaking performance trends when deployed on the Jetson AGX Orin.}
  \label{fig:nuscenes}
\end{figure}
\noindent in the literature propose various ways to overcome these challenges, such as directly processing raw unstructured points, including PointNeXt~\cite{qian2022pointnext}, WaffleIron~\cite{puy2023waffleiron} and PTv3~\cite{wu2024ptv3}. But to maintain spatial relationships, these methods often rely on time-consuming operations, such as k-nearest neighbor search, farthest point sampling, or serialization for neighbor mappings. In contrast, Superpoint Transformer~\cite{robert2023superpointtransformer} achieves significant input reduction by leveraging local similarity in dense point clouds, greatly improving efficiency. However, sparse outdoor point clouds lack sufficient local similarity, making such approaches less effective. Other methods project the point cloud into intermediate grids, such as range images, including SalsaNext~\cite{cortinhal2020salsanext}, CENet~\cite{cheng2022cenet}, RangeFormer~\cite{kong2023rangeformer} and FRNet~\cite{xu2025frnet}, to benefit from highly optimized 2D CNNs. This improves efficiency but often at the cost of lower accuracy, as spatial information can be distorted during the projection. Alternatively, Minkowski~\cite{choy2019minkowski} and Cylinder3D~\cite{zhu2021cylindrical} exploit the sparsity in the point cloud by quantizing the points into voxels and applying 3D convolution operations only on the sparse voxels, thereby discarding empty regions. However, these methods are computationally expensive and cannot be executed in real-time. Other approaches such as SPVCNN~\cite{tang2020spvnas}, RPVNet~\cite{Xu2021rpvnet} and 2DPASS~\cite{Yan20222dpass} fuse points from different views or representations to complement the information and improve predictions, but this makes them too cumbersome for mobile applications.

A recent study~\cite{abouhaidar2024realtimesemseg} systematically evaluates various 3D semantic segmentation methodologies and assesses their performance and capability for real-time deployment on resource-constrained NVIDIA Jetson embedded systems. Their findings highlight the need for substantial network redesigns to optimize performance on embedded platforms. Additionally, the authors emphasize the critical role of pre-processing, which includes the CPU-based operations required to prepare input scans prior to network inference on the GPU. This aspect is often overlooked in state-of-the-art methods but may be more critical than inference time itself.

To this end, we propose HARP-NeXt, a high-speed, accurate, and deployable network for real-time LiDAR semantic segmentation. We enhance predictions by fusing 2D range images with 3D points in a multi-scale manner to complement feature learning. The core of our backbone is Conv-SE-NeXt, a module we design to incorporate depth-wise separable convolutions that decompose convolutions into depth-wise and pointwise operations, significantly reducing the number of parameters, and to integrate a squeeze-and-excitation~\cite{hu2018squeezeexcitation} mechanism designed to adaptively recalibrate feature responses by weighing the importance of each channel, enabling the network to focus on the most informative features while optimizing both efficiency and performance. We summarize our contributions as follows:
\begin{itemize}
    \item We present a novel pre-processing workflow that leverages GPU parallel processing to accelerate data preparation, reduce CPU load, mitigate data transfer bottlenecks, and minimize the overhead prior to inference;
    \item We propose Conv-SE-NeXt, a feature extraction module that effectively captures patterns without relying on deep layer stacking, thus reducing computational cost while still outperforming traditional building blocks;
    \item We introduce a multi-scale range-point fusion backbone to enhance the representation of spatial features across various levels and resolutions, improving the model's ability to capture both fine-grained details and broader contextual information; 
    \item The proposed HARP-NeXt achieves the best trade-off between speed and accuracy on the nuScenes~\cite{caesar2020nuscenes} and SemanticKITTI~\cite{behley2019semantickitti} benchmarks, evaluated on both the NVIDIA RTX4090 GPU and the NVIDIA Jetson AGX Orin embedded system.
\end{itemize}

\section{RELATED WORK}
\label{sec:relatedwork}

LiDAR semantic segmentation provides fine-grained semantics of a scene by labeling or classifying raw 3D points captured from the system's sensor. Related methods are typically categorized into:

\textbf{Point-based methods.}
These methods operate directly on raw 3D points, preserving full geometric information without relying on other representations. PointNeXt~\cite{qian2022pointnext} applied shared multi-layer perceptrons (MLPs) and global pooling but struggled with capturing local context, limiting its performance in complex scenes. WaffleIron~\cite{puy2023waffleiron} and PTv3~\cite{wu2024ptv3} introduced local feature aggregation techniques to improve accuracy, but high computational costs make them less suitable for real-time autonomous driving applications, particularly when processing large-scale LiDAR point clouds.

\textbf{Projection-based methods.} SalsaNext~\cite{cortinhal2020salsanext}, CENet~\cite{cheng2022cenet}, and FRNet~\cite{xu2025frnet} stem from the image segmentation domain, particularly leveraging CNN architectures designed for segmenting RGB images, and are adapted for processing 3D point clouds. By projecting point cloud data onto intermediate grids, they process information entirely on 2D feature maps, which generally improves computational efficiency. However, they often fall behind other methods in terms of accuracy due to the loss of spatial information during projection.

\textbf{Sparse convolution-based methods.} Such methods have gained significant popularity for processing point clouds due to their ability to efficiently handle the sparsity and varying density of outdoor LiDAR data. By focusing computations only on occupied regions and discarding empty ones, Minkowski~\cite{choy2019minkowski} and Cylinder3D~\cite{zhu2021cylindrical} reduce both memory consumption and computational costs, while still maintaining high prediction accuracy. However, despite these advantages, sparse convolution-based methods are still not suitable for real-time applications, especially on embedded platforms.

\textbf{Fusion-based methods.} These methods improve semantic segmentation by combining different data representations. SPVCNN~\cite{tang2020spvnas} and RPVNet~\cite{ Xu2021rpvnet} fuse multiple views from a single input LiDAR scanner, while 2DPASS~\cite{Yan20222dpass} combines LiDAR and camera data. By doing so, they leverage the unique advantages of each data representation, leading to better segmentation results. However, processing multiple data streams introduces additional computational overhead, which can impact the overall efficiency of these methods. In this work, we propose a lightweight fusion strategy that maps features between 3D points and 2D range images using vectorized indexing and sparse tensor operations, retaining rich spatial context, but without sacrificing efficiency.
\section{METHODOLOGY}
\label{sec:methodology}

\subsection{Pre-processing}
\label{subsec:preprocessing}
\begin{figure}[t]
  \centering
    \includegraphics[width=0.4\textwidth]{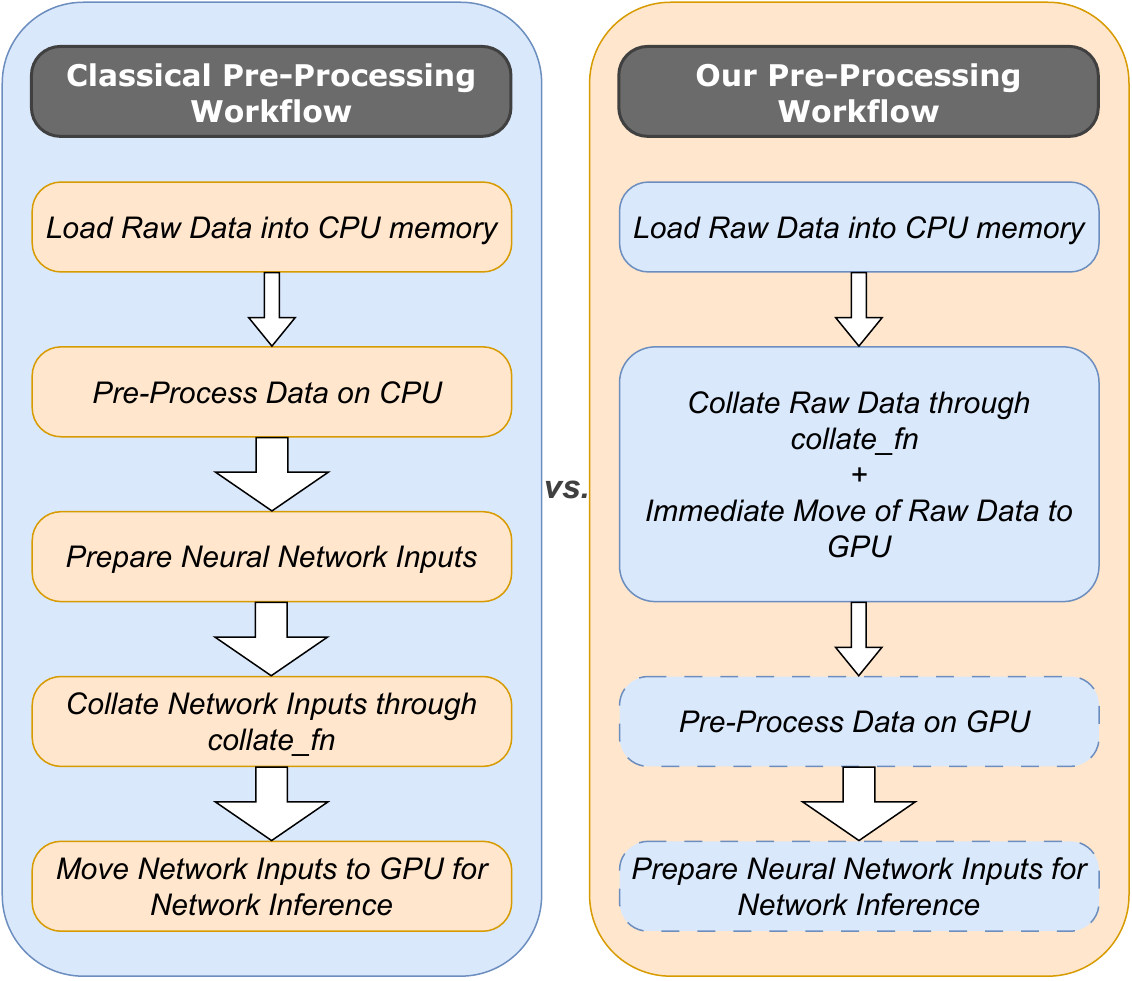}
    \caption{Pre-processing workflows. Dashed blocks indicate on GPU processes; arrow thickness reflects data bandwidth.} 
    \label{fig:preprocessing}
\end{figure}

To obtain the range image, we follow~\cite{cortinhal2020salsanext, cheng2022cenet, xu2025frnet} by first applying a spherical projection of the point cloud, converting each point $p_i = (x_i, y_i, z_i)$ through a mapping $\mathbb{R}^3 \rightarrow \mathbb{R}^2$ to spherical coordinates, and then to image coordinates $(u, v)$:
\begin{equation} \label{eq:rangeprojection}
\begin{aligned}
\left(
\begin{aligned}
    u_i \\
    v_i
\end{aligned}
\right)
    &= \left(
\begin{aligned}
    \frac{1}{2} [1 - \arctan(y_i,x_i)\pi^{-1}] W \\
    [1 - (\arcsin(z_id_i^{-1}) + f_{up})f^{-1}]H
\end{aligned}
\right),
\end{aligned}
\end{equation}

\noindent where $(H,W)$ are predefined height and width of the range image, $f = f_{up} + f_{down}$ is the vertical Field-of-View of the sensor, and $d_i= ||p_i||_2$ is the depth of each point captured by the sensor. Many recent works adopt the range projection rather than directly processing 3D points for enhanced efficiency. However, Abou Haidar et al.~\cite{abouhaidar2024realtimesemseg} demonstrates that this projection constitutes a substantial portion (up to 83\%) of the overall execution time of such methods~\cite{cortinhal2020salsanext, cheng2022cenet, xu2025frnet}. Therefore, we propose a novel pre-processing workflow as shown in \cref{fig:preprocessing} to largely accelerate the pre-processing phase prior to network inference. The classical pre-processing workflow runs entirely on the CPU and transfers only the prepared network inputs to the GPU for inference. Our approach leverages the GPU for data pre-processing, reducing CPU load and minimizing data transfer bottlenecks (network inputs have higher bandwidth than raw data due to additional features). By moving raw data to the GPU earlier, we enable parallel pre-processing and improve data throughput. Our workflow proves to be advantageous for real-time applications.
\subsection{HARP-NeXt Architecture}
\label{subsec:architecture}
\begin{figure}[t]
  \centering
    \includegraphics[width=0.485\textwidth]{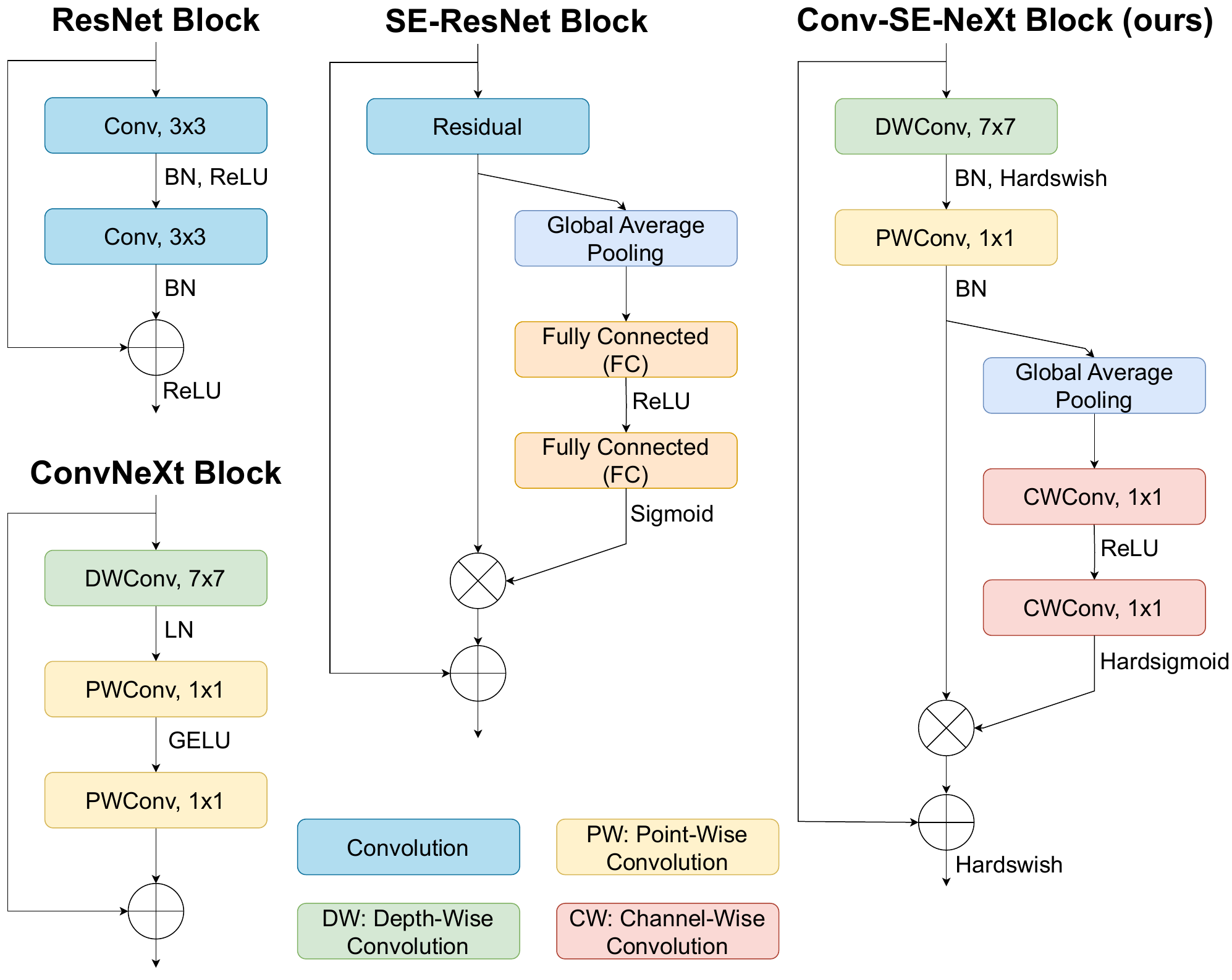}
    \caption{Design structure of ResNet~\cite{he2016resnet}, ConvNeXt~\cite{liu2022convnext}, SE-ResNet~\cite{hu2018squeezeexcitation} and \textbf{Conv-SE-NeXt} blocks.}
    \label{fig:buildingblocks}
\end{figure}

\begin{figure*}[t]
  \centering
    \includegraphics[width=\textwidth]{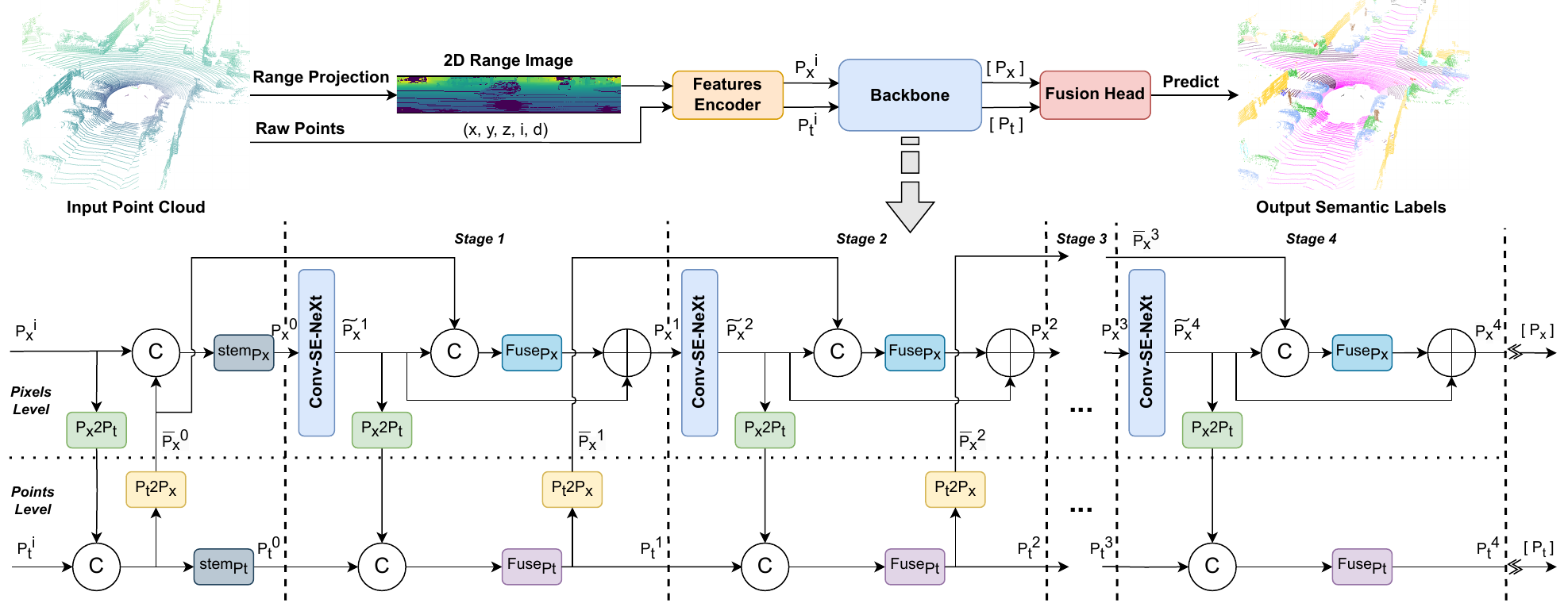}
    \caption{\textbf{HARP-NeXt} architecture consists of: 1) a Features Encoder that embeds initial 2D and 3D features, 2) a Backbone that employs a single Conv-SE-NeXt feature extractor per network stage, and fuses pixel and point features at multi-scale via efficient mappings $\mathcal{P}_{t}2\mathcal{P}_{x} : \mathcal{P}_{x} = \mathcal{M}(\mathcal{P}_{t})$ and $\mathcal{P}_{x}2\mathcal{P}_{t} : \mathcal{P}_{t} = \mathcal{M}^{-1}(\mathcal{P}_{x})$ to hierarchically refine them, and 3) a Fusion Head that combines different context-aware information from pixel and point levels to predict a label for each 3D point.}
    \label{fig:network}
\end{figure*}

For efficient and accurate LiDAR semantic segmentation, we propose HARP-NeXt, which leverages both 2D range images and 3D point representations to hierarchically extract and fuse pixel $(\mathcal{P}_{x})$ and point $(\mathcal{P}_{t})$ features at multiple scales enhancing the network's learning capability. To accelerate inference, our approach employs a single fundamental feature extraction block per network stage: \textbf{Conv-SE-NeXt} as seen in \cref{fig:buildingblocks}, a novel design we propose to effectively extract low-level features without requiring multiple stacked blocks per network stage, and to ensure that each stage remains lightweight and efficient (see \cref{fig:network}). 

\textbf{Conv-SE-NeXt Block.} \cref{fig:buildingblocks} shows our design inspiration from ResNet~\cite{he2016resnet}, ConvNeXt~\cite{liu2022convnext}, and SE-ResNet~\cite{hu2018squeezeexcitation} blocks, but with a goal of accelerating inference through maximizing computational efficiency and feature extraction capabilities without the need to stack multiple blocks at each network stage. Hence, we first employ depth-wise separable convolution to decouple the spatial and channel-wise convolutions, significantly reducing the number of parameters and computation. The depth-wise convolution captures spatial patterns with a large receptive field (3$\times$3, 5$\times$5 or 7$\times$7 kernel) allowing broader context capture without additional layers. Then a (1$\times$1 kernel) point-wise convolution is used to aggregate features across channels:
\begin{equation}
    \mathbf{y}_{dw} = \sigma(\mathcal{N}(\mathbf{W}_{dw} * \mathbf{x})),
\end{equation}
\begin{equation}
    \mathbf{y}_{pw} = \mathcal{N}(\mathbf{W}_{pw} * \mathbf{y}_{dw}),
\end{equation}
where
\(\mathbf{W}_{dw}\) and \(\mathbf{W}_{pw}\) represent the depthwise and pointwise convolution kernels, respectively, $*$ is the convolution operation. \(\mathcal{N}(\cdot)\) denotes Batch Normalization, and \(\sigma(\cdot)\) represents the Hardswish activation function, selected for its efficient computation and smooth nonlinearity, as introduced in MobileNetV3~\cite{howard2019mobilenetv3} for mobile and embedded systems:
\begin{equation}
    \sigma(x) = x \cdot \text{ReLU6}(x + 3) / 6,
\end{equation}
where \(\text{ReLU6}(x)\) is the activation function capped at 6.
\noindent We then apply the Squeeze-and-Excitation (SE) module as a weight to the features $\mathbf{y}_{pw}$  to learn an attention mechanism over the channel dimension. But instead of using Fully Connected (FC) layers as in~\cite{hu2018squeezeexcitation} to compute this attention, we opted for 1x1 channel-wise convolutions, which are computationally more efficient and maintain the model's expressiveness. A channel descriptor for each feature map is first obtained by global average pooling (GAP):
\begin{equation}
    \mathbf{z}_c = \text{GAP}(\mathbf{y}_{pw}) = \frac{1}{H \times W} \sum_{i=1}^{H} \sum_{j=1}^{W} \mathbf{y}_{pw}(i, j).
\end{equation}
Then, we capture channel dependencies and generate an attention weight for each channel by:
\begin{equation}
    \mathbf{s}_c = \sigma_2(\mathbf{W}_2 \sigma_1(\mathbf{W}_1 \mathbf{z}_c)),
\end{equation}
where
\(\mathbf{W}_1\) and \(\mathbf{W}_2\) are learned convolutional weights,
\(\sigma_1(\cdot)\) is a ReLU activation function, and
\(\sigma_2(\cdot)\) is a Hardsigmoid activation function, which was chosen over sigmoid for its efficiency. Finally, we apply the SE weight $\mathbf{s}_c$ to the feature map $\mathbf{y}_{pw}$ by performing an element-wise multiplication:
\begin{equation}
    \tilde{\mathbf{y}}_c = \mathbf{y}_{pw} \cdot \mathbf{s}_c
\end{equation}
By doing that, the SE module enhances the model's ability to focus on important channels, improving feature selectivity without the need for additional layers. We finally add a skip connection to help mitigate the vanishing gradient problem:
\begin{equation}
    \mathbf{z} = \tilde{\mathbf{y}}_c + \mathbf{x}
\end{equation}
\hspace{-0.13cm} \cite{uzen2022dseb} also proposed a depth-wise Squeeze and Excitation block (DSEB), to calculate the squeeze weight matrix $\mathbf{s}_c = \frac{1}{1 + e^{-\mathbf{y}_{pw}}}$ from the point-wise convolution and apply it directly to the skip connection: $\mathbf{z} = \mathbf{x} \cdot \mathbf{s}_c$ as in~\cite{mehmet2024dsenet}, while using different nonlinearities. MobileNetV3~\cite{howard2019mobilenetv3} also integrates SE blocks selectively into inverted residual blocks. We demonstrate the superiority of Conv-SE-NeXt block over DSEB and MobileNetV3 blocks in the ablation study.

\textbf{Features Encoder.} This takes as input the raw point cloud  $\mathcal{P} \in \mathbb{R}^{N \times C}$, where $N$ is the number of points, and $C$ represents the Cartesian spatial coordinates along with the LiDAR intensity and depth of each point, given as $(x, y, z, i, d)$. It also takes as input the range image $\mathcal{I} \in \mathbb{R}^{H \times W}$, and the set of $K$ points projected to the same pixel area $(u,v)$ in \eqref{eq:rangeprojection} clustered as $\widetilde{\mathcal{P}} = \{\widetilde{\mathcal{P}}_{1}, \widetilde{\mathcal{P}}_{2}, \dots, \widetilde{\mathcal{P}}_{K}\}$ where we apply an average pooling function to obtain a representation $\widetilde{\mathcal{P}}_{m}$ for each pixel. Finally, the initial per-point features are extracted by a series of MLPs:
\begin{equation}
\label{eq:featuresencoderpoints}
    \mathcal{P}_{t}^{i} = \text{MLP}([\mathcal{P};\mathcal{P} - \widetilde{\mathcal{P}}_{m}]), 
\end{equation}
where $\left[ \cdot \; ; \; \cdot \right]$ denotes features concatenation. Then for every pixel in the range image $\mathcal{I}$ we apply a max pooling function over the corresponding clustered points to get pixel features:
\begin{equation}
\label{eq:featuresencoderpixels}
    \mathcal{P}_{x}^{i} = \text{MLP}(\text{MAX}(\mathcal{P}_{t}^{i})).
\end{equation}

\textbf{Multi-scale Range-Point Fusion Backbone.} We employ a backbone with four main stages to hierarchically fuse features from both pixel and point levels. We utilize our \textbf{Conv-SE-NeXt} module to effectively extract the pixel features while maintaining computational efficiency. At each stage, an efficient mapping between pixels and points is incorporated to refine features at both levels. Let $\mathcal{P}_{t}2\mathcal{P}_{x}$ denotes a mapping function $\mathcal{P}_{x} = \mathcal{M} \left( \mathcal{P}_{t} \right)$ from point to pixel features: $\mathbb{R}^{N \times C} \rightarrow \mathbb{R}^{H \times W \times C}$, and $\mathcal{P}_{x}2\mathcal{P}_{t}$ an inverse mapping $\mathcal{P}_{t} = \mathcal{M}^{-1} \left( \mathcal{P}_{x} \right)$ from pixel to point features: $\mathbb{R}^{H \times W \times C} \rightarrow \mathbb{R}^{N \times C}$. Prior to the network stages, we use these mapping functions to concatenate features from both point and pixel levels, which are then encoded by the stem layers as shown in \cref{fig:network}. The pixel stem $(\text{stem}_{\mathcal{P}_{x}})$ consists of three sequential convolutional layers, each followed by batch normalization and Hardswish nonlinearity, whereas the point stem $(\text{stem}_{\mathcal{P}_{t}})$ comprises a single linear layer followed by batch normalization and Hardswish activation. In each stage, we then apply multi-scale fusion at the pixel level where we first concatenate pixel features from different scales using bilinear interpolation and fuse them with:
\begin{equation}
    \mathcal{F}_x^n = \text{Conv}\left( \tilde{\mathcal{P}}_x^n \parallel \text{Interpolate}(\bar{\mathcal{P}}_x^{n-1}) \right),
\end{equation}
where $\parallel$ denotes a channel-wise concatenation, $\tilde{\mathcal{P}}_x^n$ is the extracted features at stage $n$ from the \textbf{Conv-SE-NeXt} module, $\bar{\mathcal{P}}_x^{n-1}$ is the mapped pixel features from points in the previous stage $n-1$, and $\text{Conv}\left( \cdot \right)$ represents a convolution layer with batch normalization and Hardswish activation. Then we calculate an attention map to weight the fused features obtained from multiple scales:
\begin{equation}
    \mathcal{A} = \sigma \left( h(\mathcal{F}_x^n, \theta) \right),
\end{equation}
where $\sigma$ is a sigmoid function, $\text{h}\left( \cdot \right)$ is a linear function with trainable parameters $\theta$. And finally we apply the attention on the features and refine pixel features in a residual-attentive manner:
\begin{equation}
    \mathcal{P}_x^n = \tilde{\mathcal{P}}_x^n +  \mathcal{A}  \odot \mathcal{F}_x^n, 
\end{equation}
where $\odot$ denotes element-wise multiplication. In each stage, \( \tilde{\mathcal{P}}_x^n \) is also used to refine the points' features by combining them with those from the previous stage: 
\begin{equation}
    \mathcal{P}_t^n = \sigma \left( h \left( \mathcal{M}^{-1}(\tilde{\mathcal{P}}_x^n) \parallel (\mathcal{P}_t^{n-1}), \theta \right) \right).
\end{equation}

\textbf{Fusion Head.} Inspired by CENet~\cite{cheng2022cenet} and FRNet~\cite{xu2025frnet}, we leverage multi-stage features to capture diverse context-aware information, enhancing the network's predictive capability. Hence, we aggregate pixel and point features with different receptive fields from each network stage, along with $\mathcal{P}_x^0$ and $\mathcal{P}_t^0$ which encode the initial descriptors. In the Fusion Head, we fuse the aggregated features separately to refine each of pixel-level and point-level feature representation:
\begin{equation}
    \mathcal{P}_x^f = \text{Conv}\left([\mathcal{P}_{x}]=[\mathcal{P}_x^0, \mathcal{P}_x^1, \dots, \mathcal{P}_x^4]\right),
\end{equation}
\begin{equation}
    \mathcal{P}_t^f = \text{MLP}\left([\mathcal{P}_{t}]=[\mathcal{P}_t^0, \mathcal{P}_t^1, \dots, \mathcal{P}_t^4]\right).
\end{equation}
$\mathcal{P}_x^f$ contains global information from the clustered points within the corresponding pixel of the range image, while  $\mathcal{P}_t^f$ contains fine-grained local information in the 3D space. Hence, we fuse them together to help extract features from global to local and obtain the logit-level point features:
\begin{equation}
    \mathcal{P}_t^{logit} = \text{MLP}\left(\mathcal{M}^{-1}(\mathcal{P}_x^f) + \mathcal{P}_t^f\right).
\end{equation}
A linear head is then used on $\mathcal{P}_t^{logit}$ to predict the final semantic scores for each 3D point in the point cloud.

\subsection{Loss Function}
To enhance semantic learning, we apply supervision at both pixels and points levels. As in~\cite{cheng2022cenet}, we incorporate multiple auxiliary loss heads at each network stage to refine feature maps at different resolutions, enhancing pixel-level learning with dedicated loss functions. Typically, a pixel in $\mathcal{I}$ encompasses multiple projected points, which may belong to different categories. Inspired by~\cite{xu2025frnet}, we assign pseudo-labels to each pixel by selecting the most frequent category among the projected points within each pixel. We also employ a point-level loss function to supervise the final semantic scores. The overall loss function is defined as:
\begin{equation}
    \mathcal{L} = \mathcal{L}_{ce}^{pt} + \lambda \left(\mathcal{L}_{ce}^{px} + \beta\mathcal{L}_{ls}^{px} + \mathcal{L}_{bd}^{px}\right).
\end{equation}
Here, $\mathcal{L}_{ce}^{pt}$ and $\mathcal{L}_{ce}^{px}$ denote the point-level and pixel-level cross-entropy loss functions respectively. $\mathcal{L}_{ls}^{px}$ represents the pixel-level Lovász-Softmax loss~\cite{berman2018lovaszsoftmaxloss}, and $\mathcal{L}_{bd}^{px}$ is the pixel-level boundary loss~\cite{bokhovkin2019boundaryloss}. The hyperparameter $\lambda$ balances the pixel-level supervision and is set to 1.0, while $\beta$ controls the weight of the Lovász-Softmax loss and is set to 1.5.

\section{EXPERIMENTAL RESULTS}
\label{sec:experimentalresults}
We evaluate the performance of the proposed HARP-NeXt on nuScenes~\cite{caesar2020nuscenes} and SemanticKITTI~\cite{behley2019semantickitti} benchmarks.

\textbf{nuScenes.} An outdoor dataset that encompasses a comprehensive autonomous vehicle sensor suite including a Velodyne HDL-32E LiDAR, offering a complete 360\degree\ Field-of-View (FoV) while driving in Boston and Singapore, two cities very well known for their dense traffic and challenging driving situations. 

\textbf{SemanticKITTI.} This dataset showcases urban outdoor environments captured with a Velodyne HDL-64E LiDAR covering the full 360\degree\ FoV. We follow the same division of its 22 sequences, and only work with sequences 00 to 10 designated for training except for the 8$^{\text{th}}$ used for validation. 

\textbf{Evaluation metrics.} We adopt the mean Intersection-over-Union (mIoU) over all evaluation classes, formulated as:
\begin{equation} 
 mIoU = \frac{1}{C} \displaystyle\sum_{c=1}^{C}\frac{TP_c}{TP_c + FP_c + FN_c}
 \label{eq:mIoU}
\end{equation}
where $TP_c$, $FP_c$, and $FN_c$ refer respectively to the number of true positive, false positive, and false negative predictions for class $c$, and $C$ the total number of classes. We also assess efficiency by measuring the pre-processing and inference time, which together constitute the Total Runtime of a network, along with the total number of parameters, multiply-accumulate operations (MACs), and peak GPU memory allocated. These metrics are profiled on the same set of 1,000 scans from each dataset, and we report the mean values.

\subsection{Experimental Setup}
\label{subsec:experimentalsetup}
For a fair comparison with state-of-the-art methods, all models are reproduced and retrained without additional training data, as in~\cite{wu2024ptv3}, or test-time augmentation (TTA), as in~\cite{puy2023waffleiron}. Omitting TTA, which applies augmentations such as flipping, scaling, or employing model ensembles at inference time, ensures that reported results in \cref{tab:results} reflect the models' inherent capabilities rather than post-processing enhancements.

\textbf{HARP-NeXt Setup.} We project the point cloud onto range images with a resolution of $32 \times 480$ for nuScenes and $64 \times 512$ for SemanticKITTI. The input to the Features Encoder consists of 5 channels: the Cartesian coordinates, intensity, and depth. The latter incorporates cluster center information and processes these inputs through MLPs with the following channel configurations: 64, 128, 256, and 256 for point features in~\eqref{eq:featuresencoderpoints}, and 16 for pixel features in~\eqref{eq:featuresencoderpixels}. The Backbone having 4 stages with strides (1, 2, 2, 2) and dilations (1, 1, 1, 1) takes 16 input channels for pixel features and 256 for point features, and generates a feature map with dimensions [128, 128, 128, 128] at both the pixel and point levels. We set the kernel size for the depth-wise convolution in \textbf{Conv-SE-NeXt} to $3\times3$ on nuScenes and $7\times7$ on SemanticKITTI. Finally, the Fusion Head reduces 128 input channels to 64, followed by a segmentation layer for classification.

\begin{table*}[h]
    \centering
    \caption{Models' efficiency and performance trade-off on nuScenes and SemanticKITTI val sets. Total Runtime is the sum of pre-processing and inference. The \textbf{best} and \underline{second best} scores are highlighted in \textbf{bold} and \underline{underline} for each dataset.}
    \label{tab:results}
    \resizebox{0.99\textwidth}{!}{
    \begin{tabular}{c|r|c|c|c|c c|c c|c|c|c}
        \toprule
         & \multirow{2}{*}{\textbf{Methods}} & \multirow{2}{*}{\textbf{Year}} & \multirow{2}{*}{\textbf{Category}} & \multirow{2}{*}{\textbf{mIoU (\%)}}  & \multicolumn{2}{c|}{\textbf{Pre-Processing (ms)}} & \multicolumn{2}{c|}{\textbf{Total Runtime (ms)}} & \multirow{2}{*}{\textbf{\#Params (M)}} & \multirow{2}{*}{\textbf{\#MACs (G)}} & \textbf{GPU} \\
        & & & & & \textbf{RTX4090} & \textbf{AGX Orin} & \textbf{RTX4090} & \textbf{AGX Orin} & & &  \textbf{Memory (MB)} \\
        \midrule
        \multirow{10}{*}{\rotatebox{90}{\textbf{nuScenes~\cite{caesar2020nuscenes}}}} & WaffleIron~\cite{puy2023waffleiron} & 2023 & \multirow{2}{*}{Point-based} & 76.1 & 64 & 114 & 111 & 527 & 6.8 & 122.4 & 750 \\
        & PTv3~\cite{wu2024ptv3} & 2024 &  & \textbf{78.4} & 198 & 736 & 241 & 872 & 15.3 & 96.9 & 346 \\
        \cmidrule{2-12}
        & SalsaNext~\cite{cortinhal2020salsanext} & 2020 & \multirow{3}{*}{Projection-based} & 68.2 & \underline{8} & 26 & \underline{13} & \textbf{51} & \underline{6.7} & 31.4 & \underline{329} \\
        & CENet~\cite{cheng2022cenet} & 2022 & & 73.3 & 11 & 58 & 16 & 97 & 6.8 & 54.4 & \textbf{269} \\
        & FRNet~\cite{xu2025frnet} & 2025 &  & 75.1 & 70 & 252 & 82 & 383 & 10.0 & 98.8 & 521 \\
        \cmidrule{2-12}
        & Minkowski~\cite{choy2019minkowski} & 2019 & \multirow{2}{*}{Sparse Conv-based} & 73.5 & \underline{8} & \underline{22} & 47 & 146 & 21.7 & \textbf{22.2} & 351 \\
        & Cylinder3D~\cite{zhu2021cylindrical} & 2021 & & 76.1 & 136 & - & 176 & - & 55.9 & 41.7 & 1,421 \\
        \cmidrule{2-12}
        & SPVCNN~\cite{tang2020spvnas} & 2020 & \multirow{2}{*}{Fusion-based} & 72.6 & \underline{8} & 23 & 57 & 169 & 21.8 & \underline{23.1} & 383 \\
        & \textbf{HARP-NeXt (ours)} & 2025 &  & \underline{77.1} & \textbf{3} & \textbf{10} & \textbf{10} & \underline{71} & \textbf{5.4} & 79.1 & 448 \\
        \midrule
        \midrule
        \multirow{9}{*}{\rotatebox{90}{\textbf{SemanticKITTI~\cite{behley2019semantickitti}}}} & WaffleIron~\cite{puy2023waffleiron} & 2023 & \multirow{1}{*}{Point-based} & \underline{65.8} & 232 & 381 & 370 & 1,847 & 6.8 & 457.1 & 2,384 \\
        \cmidrule{2-12}
        & SalsaNext~\cite{cortinhal2020salsanext} & 2020 & \multirow{3}{*}{Projection-based} & 55.9 & 28 & 58 & \underline{36} & \textbf{109} & \underline{6.7} & \textbf{62.8} & \textbf{498} \\
        & CENet~\cite{cheng2022cenet} & 2022 &  & 62.6 & 31 & 62 & 58 & 165 & 6.8 & 435.2 & 968 \\
        & FRNet~\cite{xu2025frnet} & 2025 &  & \textbf{66.0} & 67 & 194 & 86 & 394 & 10.0 & 218.6 & 999 \\
        \cmidrule{2-12}
        & Minkowski~\cite{choy2019minkowski} & 2019 & \multirow{2}{*}{Sparse Conv-based} & 64.3 & \underline{22} & 45 & 71 & 211 & 21.7 & \underline{113.9} & \underline{542} \\
        & Cylinder3D~\cite{zhu2021cylindrical} & 2021 & & 63.2 & 261 & - & 309 & - & 55.9 & 140.6 & 1,629 \\
        \cmidrule{2-12}
        & SPVCNN~\cite{tang2020spvnas} & 2020 & \multirow{2}{*}{Fusion-based} & 65.3 & 26 & \underline{43} & 85 & 252 & 21.8 & 118.6 & 657 \\
        & \textbf{HARP-NeXt (ours)} & 2025 & & 65.1 & \textbf{3} & \textbf{13} & \textbf{13} & \underline{120} & \textbf{5.4} & 174 & 872 \\
        \bottomrule
    \end{tabular}}
\end{table*}

\textbf{Training Protocol.} We standardize the training for all methods, and train on a single NVIDIA GeForce RTX4090 GPU using the AdamW optimizer for 80 epochs on nuScenes~\cite{caesar2020nuscenes} and SemanticKITTI~\cite{behley2019semantickitti}, with a weight decay of 0.003 and a batch size of 4. The learning rate is scheduled with a linear warmup from 0 to 0.001 over the first 4 epochs, followed by cosine annealing to decay it to $10^{-5}$ by the final epoch. To prevent overfitting, we apply classical point cloud augmentations such as rotation around the z-axis, flipping along the x or y axes, and random re-scaling. Inspired by~\cite{xu2025frnet}, we also apply random horizontal and vertical point cloud mixing based on pitch and yaw angles, along with interpolation on empty pixels in the range image (when applicable). Following~\cite{puy2023waffleiron, Xu2021rpvnet, Yan20222dpass}, we implement instance cutmix and polarmix on SemanticKITTI, targeting rare classes such as bicycle, motorcycle, person, and bicyclist. For each selected class, up to 40 instances are randomly chosen, transformed (e.g., rotation, flipping, re-scaling) and placed randomly on roads, parking areas, or sidewalks to enhance segmentation.



        



\subsection{Results}
\label{subsec:results}
As shown in \cref{tab:results}, we compare the proposed HARP-NeXt with state-of-the-art methods. HARP-NeXt demonstrates superior efficiency and competitive accuracy across both nuScenes and SemanticKITTI. It achieves the second-highest mIoU (77.1\%) on nuScenes, outperforming all projection, sparse convolution, and fusion-based methods and is comparable to the first-ranking PTv3~\cite{wu2024ptv3}, while maintaining the fastest Total Runtime on the RTX4090 and the second-fastest on the AGX Orin. Unlike point-based methods, which suffer from high computational costs (e.g., PTv3 requires on average 241 ms on RTX4090 and 872 ms on AGX Orin for execution), HARP-NeXt achieves high accuracy while running at 10 ms on the RTX4090 and 71 ms on the AGX Orin. It provides a significant speed advantage, with an execution 7 to 24$\times$ faster than the closest accurate methods. We observe the same trend on SemanticKITTI, where HARP-NeXt achieves comparable accuracy (65.1\% mIoU) to state-of-the-art methods while significantly outperforming them in efficiency. Furthermore, it maintains the lowest parameter count (5.4M) and moderate computational complexity in terms of MACs and GPU memory consumption, making it an optimal choice for resource-constrained platforms. 

\begin{figure}[H]
  \centering
    \includegraphics[width=0.455\textwidth]{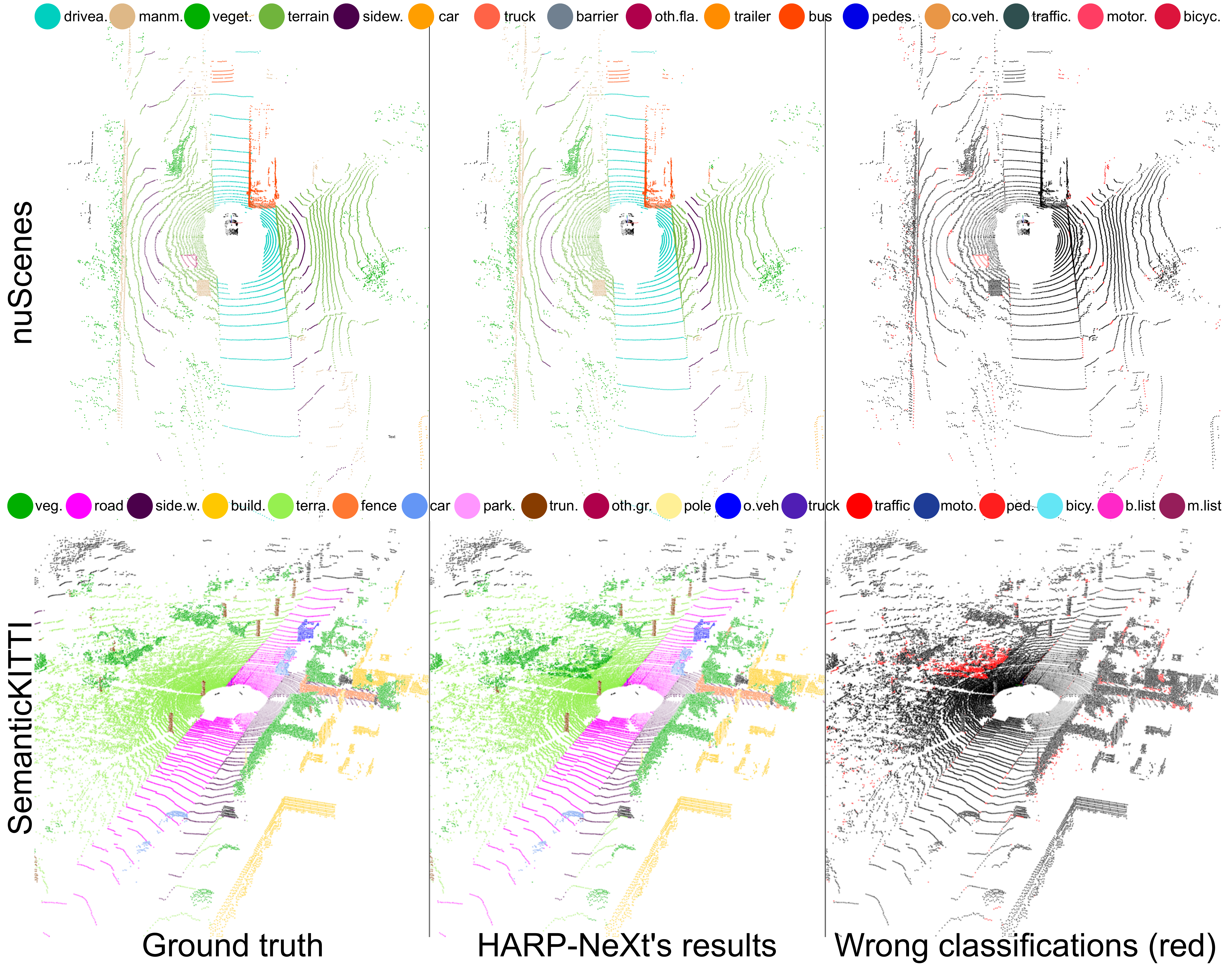}
    \caption{HARP-NeXt's results compared to the ground truth.}
    \label{fig:gttpfp}
\end{figure}

\begin{figure*}[htbp]
  \centering
    \includegraphics[width=0.82\textwidth]{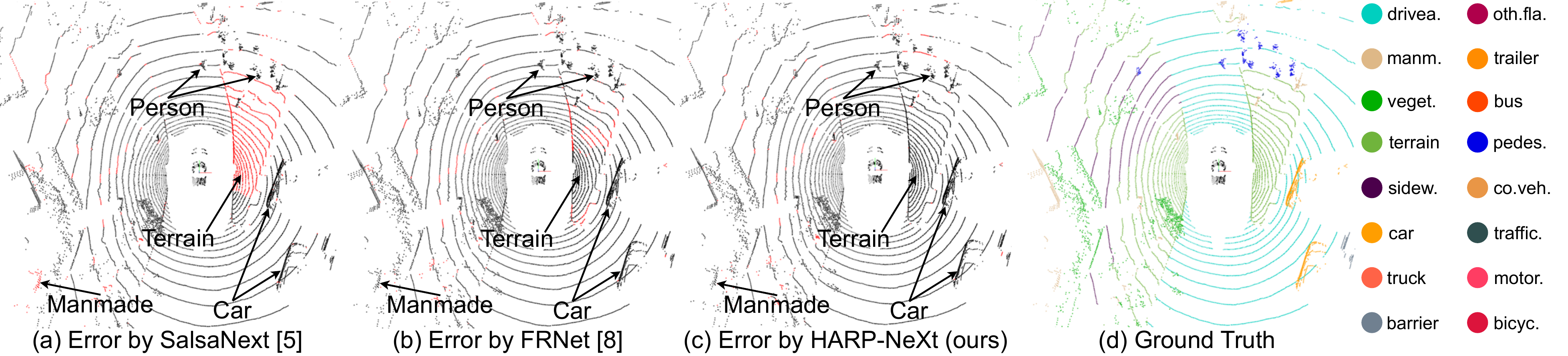}
    \caption{Error comparison on nuScenes~\cite{caesar2020nuscenes}. Wrong classifications are in red.}
    \label{fig:comparison_nuscenes}
\end{figure*}

\begin{figure*}[htbp]
  \centering
    \includegraphics[width=0.82\textwidth]{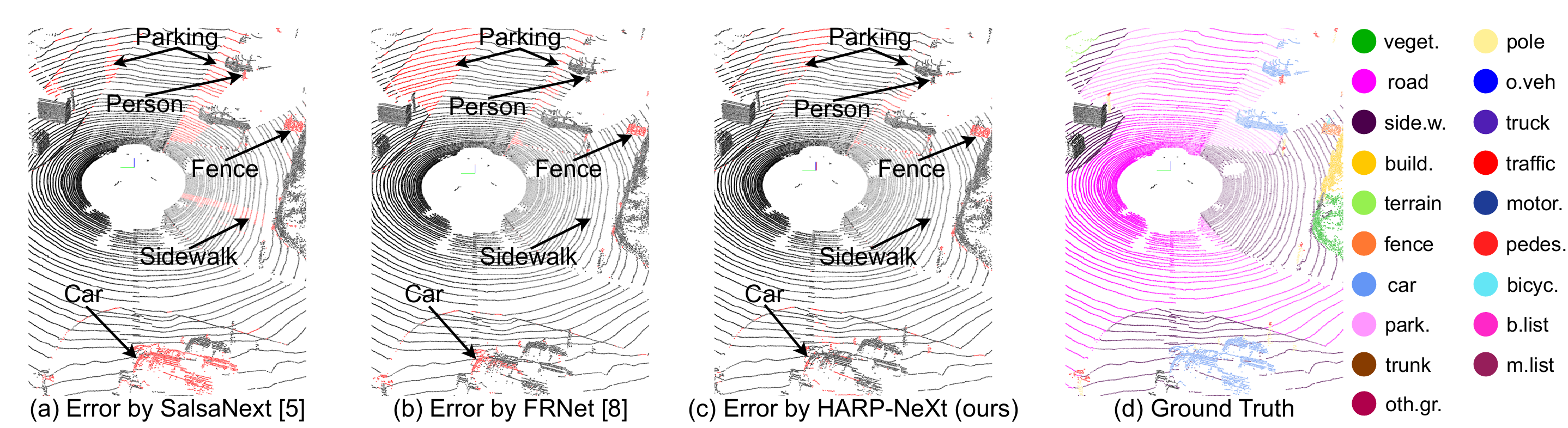}
    \caption{Error comparison on SemanticKITTI~\cite{behley2019semantickitti}. Wrong classifications are in red.}
    \label{fig:comparison_semantickitti}
\end{figure*}

\cref{tab:results} also highlights HARP-NeXt's superior pre-processing efficiency, driven by our novel workflow in~\cref{fig:preprocessing}, which outperforms all other methods. In contrast to \cite{puy2023waffleiron} and \cite{wu2024ptv3}, which rely on CPU-bound operations (we use an AMD Ryzen 9 5900X CPU) like KDTrees, space-filling curves, and point cloud serialization—tasks that involve hierarchical data structures or sequential steps, making them difficult to parallelize on a GPU—HARP-NeXt’s pre-processing presented in~\eqref{eq:rangeprojection} is GPU-executable. \cref{fig:gttpfp} presents qualitative results of our method compared to the ground truth, highlighting that misclassifications primarily occur for less critical classes (e.g., vegetation), rather than those that are more important from a safety perspective in autonomous driving. Additional qualitative results are shown in \cref{fig:comparison_nuscenes,fig:comparison_semantickitti},
where we compare HARP-NeXt's errors to those of other methods, highlighting the superiority of our network's predictions. We demonstrate that while SalsaNext~\cite{cortinhal2020salsanext} is efficient, it falls short in terms of accuracy. Meanwhile, FRNet~\cite{xu2025frnet}, which employs a ResNet34~\cite{he2016resnet} backbone with multiple stacked layers and improves upon RangeFormer~\cite{kong2023rangeformer} by achieving higher segmentation accuracy and up to 5$\times$ greater efficiency, still exhibits more errors than HARP-NeXt while being $3$ to $8\times$ slower. For instance, HARP-NeXt is the only method to correctly classify most points at an interaction with pedestrians on nuScenes, while SalsaNext and FRNet misclassify the terrain where pedestrians walk. On SemanticKITTI, HARP-NeXt correctly identifies the person, car, sidewalk, and the right parking spot, whereas SalsaNext, which is the closest in efficiency, misclassifies all of them, and FRNet shows comparable errors across classes. \Cref{tab:nuscenes_iou} presents the per-class IoU scores, where HARP-NeXt ranks first in 6 classes and second in 4, showing a strong performance across a wide
\vspace{-0.4cm} 
\begin{table*}[htbp]
\centering
\caption{Class-wise IoU on nuScenes \cite{caesar2020nuscenes} val set. The best and second best scores are in \textbf{bold} and \underline{underlined}, respectively.}
\label{tab:nuscenes_iou}
\scriptsize
\resizebox{0.85\textwidth}{!}{
\setlength{\tabcolsep}{3pt}
\begin{tabular}{r| c| c c c c c c c c c c c c c c c c}
\toprule
Method  
& \rotatebox{70}{mIoU \%}
& \rotatebox{70}{barrier} 
& \rotatebox{70}{bicycle} 
& \rotatebox{70}{bus} 
& \rotatebox{70}{car} 
& \rotatebox{70}{const. veh.} 
& \rotatebox{70}{motorcycle} 
& \rotatebox{70}{pedestrian} 
& \rotatebox{70}{traffic cone} 
& \rotatebox{70}{trailer}
& \rotatebox{70}{truck}
& \rotatebox{70}{driv. surf.}
& \rotatebox{70}{other flat}
& \rotatebox{70}{sidewalk}
& \rotatebox{70}{terrain}
& \rotatebox{70}{manmade}
& \rotatebox{70}{vegetation} \\

\midrule

WaffleIron \cite{puy2023waffleiron}
& 76.1
& \textbf{77.8}
& \underline{45.8}
& \underline{93.7}
& 85.3
& 50.2
& \textbf{81.9}
& 75.8 
& 66.2 
& \underline{68.7}
& 83.2
& 96.7 
& 72.9 
& 74.9
& 73.7 
& 86.3
& 84.0 \\

PTv3 \cite{wu2024ptv3}
& \textbf{78.4}
& 71.9
& \textbf{59.1}
& 91.0
& 90.1
& 39.1
& 70.7
& \textbf{82.7}
& \textbf{77.7}
& \textbf{84.9}
& 80.9 
& \textbf{97.8}
& 70.9 
& \underline{79.0}
& \underline{79.7}
& \textbf{91.8}
& 87.5 \\

SalsaNext \cite{cortinhal2020salsanext}
& 68.2 
& 68.1 
& 15.2 
& 76.3 
& 85.4 
& 34.0 
& 73.5 
& 71.1 
& 57.1 
& 56.4 
& 72.4 
& 95.6 
& 70.0 
& 72.6 
& 72.9 
& 85.8 
& 84.3  \\

CENet \cite{cheng2022cenet}
& 73.3
& 75.8
& 36.4
& 87.7
& 88.5
& 44.6
& 73.4
& 75.2
& 65.1
& 62.3
& 77.5 
& 96.2
& 71.8 
& 72.3
& 74.5
& 86.8
& 85.4 \\

FRNet \cite{xu2025frnet} 
& 75.1
& \underline{76.8} 
& 28.8
& 87.3
& \underline{91.3}
& 43.4 
& \underline{80.3} 
& 75.9
& 64.0
& 67.3
& 81.3 
& 97.1
& \textbf{76.3} 
& 76.7
& 76.7
& 89.8
& \underline{87.9} \\

Minkowski \cite{choy2019minkowski}
& 73.5 
& 74.5
& 39.8
& 89.7
& 91.0 
& 44.3 
& 78.7
& 77.3
& 60.4
& 56.7 
& 82.6
& 95.5 
& 68.8 
& 71.3 
& 74.4 
& 86.1 
& 84.9 \\

Cylinder3D \cite{zhu2021cylindrical}
& 76.1
& 76.4
& 40.3
& 91.2
& \textbf{93.8}
& \underline{51.3}
& 78.0
& \underline{78.9} 
& 64.9 
& 62.1
& \underline{84.4}
& 96.8 
& 71.6
& 76.4
& 75.4
& 90.5
& 87.4 \\

SPVCNN \cite{tang2020spvnas}
& 72.6 
& 73.6
& 40.8
& 88.3 
& 88.6 
& 45.6
& 78.6
& 75.4 
& 59.4 
& 54.9 
& 82.3 
& 95.4 
& 67.3 
& 70.1
& 72.9 
& 85.4 
& 83.7 \\

\textbf{HARP-NeXt (ours)}
& \underline{77.1}
& 76.2
& 37.5
& \textbf{94.2}
& 91.1
& \textbf{60.8}
& 59.5
& 75.9
& \underline{73.7}
& 65.0
& \textbf{84.8}
& \underline{97.7}
& \underline{75.9}
& \textbf{79.4}
& \textbf{83.3}
& \underline{91.0}
& \textbf{88.0} \\

\bottomrule
\end{tabular}
}
\end{table*}
\noindent range of categories (a similar trend is observed on SemanticKITTI, but not shown due to space constraints). Overall, HARP-NeXt effectively balances accuracy and efficiency, outperforming all methods in real-time feasibility, making it highly suitable for autonomous vehicles and mobile robots.

\section{ABLATION STUDY}
\label{sec:ablations}
In this section, we conduct an ablation analysis on various components and configurations within HARP-NeXt network. \cref{tab:ablations} provides insights into the performance trade-offs of different feature extraction blocks, including ResNet~\cite{he2016resnet}, ConvNeXt~\cite{liu2022convnext}, SE-ResNet~\cite{hu2018squeezeexcitation}, MobileNetV3~\cite{howard2019mobilenetv3}, and DSEB~\cite{uzen2022dseb} designs. The results demonstrate that our proposed \textbf{Conv-SE-NeXt} block consistently outperforms all other designs in terms of both accuracy and efficiency. Notably, \textbf{Conv-SE-NeXt} achieves the highest mIoU on nuScenes~\cite{caesar2020nuscenes} and SemanticKITTI~\cite{behley2019semantickitti}, surpassing the traditional blocks by a significant margin (up to 3.8\% and 6.4\% mIoU difference, respectively) while maintaining superior inference speed particularly on the Jetson AGX Orin. This highlights the robustness of \textbf{Conv-SE-NeXt} as a leading feature extraction block, demonstrating its ability to balance high performance with computational efficiency. 

Furthermore, we assess the effect of varying the number of stages in HARP-NeXt: reducing it from 4 to 3 limits the depth of feature extraction, leading to a decrease in mIoU, while offering only marginal efficiency gains. On the other hand, increasing the number of stages to 5 results in a slower inference and a reduction in accuracy. This suggests that additional stages can introduce redundancies. \cref{tab:ablations} also shows that increasing the number of blocks per stage (as in 1-2-2-1 or 2-2-2-2) with identical feature representation introduces unnecessary complexity to the network. This can lead to overfitting, resulting in suboptimal mIoU scores. Lastly, we integrate FRNet Fusion~\cite{xu2025frnet}, which fuses features within the same stage level, and it shows a close accuracy but also increases the computational cost. This is because, unlike our Multi-scale Fusion strategy which combines information from different levels of abstraction, it requires recomputing features at each stage, whereas we have them readily available from the previous stage.
\begin{table*}[htbp]
    \centering
    \caption{Ablation study of various building blocks, configurations and fusion strategies in HARP-NeXt architecture.}
    \label{tab:ablations}
    \resizebox{0.85\textwidth}{!}{
    \begin{tabular}{r|c|c|c|ccc|ccc}
        \toprule
        \multirow{3}{*}{\textbf{Building Block}} & \multirow{3}{*}{\textbf{\#Stages}} & \multirow{3}{*}{\textbf{\#Blocks/Stage}} & \multirow{3}{*}{\textbf{Fusion}} & \multicolumn{3}{c|}{\textbf{nuScenes~\cite{caesar2020nuscenes}}} & \multicolumn{3}{c}{\textbf{SemanticKITTI~\cite{behley2019semantickitti}}} \\
        \cline{5-10}
        & & & & \multirow{2}{*}{\textbf{mIoU}} & \multicolumn{2}{c|}{\textbf{Inference (ms)}} & \multirow{2}{*}{\textbf{mIoU}} & \multicolumn{2}{c}{\textbf{Inference (ms)}} \\
        \cline{6-7} \cline{9-10}
        & & & & & \textbf{RTX} & \textbf{Orin} & & \textbf{RTX} & \textbf{Orin} \\
        \midrule
        ResNet Block~\cite{he2016resnet} & \multirow{5}{*}{4} & \multirow{5}{*}{1-1-1-1} & \multirow{5}{*}{Multi-scale (ours)} & 74.2 & 7 & 69 & 61.4 & 10 & 115 \\
        ConvNeXt Block~\cite{liu2022convnext} &  &  & & 73.3 & 7 & 66 & 59.5 & 10 & 110 \\
        SE-ResNet Block~\cite{hu2018squeezeexcitation} &  &  & & 74.2 & 8 & 72 & 58.8 & 11 & 109 \\
        MobileNetV3 Block~\cite{howard2019mobilenetv3} &  &  & & 76.9 & 8 & 73 & 60.4 & 11 & 120 \\
        DSEB Block~\cite{uzen2022dseb} &  &  & & 74.8 & 8 & 68 & 58.7 & 10 & 139 \\
        \midrule
        \multirow{2}{*}{Conv-SE-NeXt (Ours)} & 3 & \multirow{2}{*}{1-1-1-1} & \multirow{2}{*}{Multi-scale (ours)} & 71.1 & 6 & 57 & 56.8 & 9 & 92 \\
        & 5 & & & 75.4 & 9 & 75 & 59.1 & 12 & 133 \\
        \midrule
        \multirow{2}{*}{Conv-SE-NeXt (Ours)} & \multirow{2}{*}{4} & 1-2-2-1 & \multirow{2}{*}{Multi-scale (ours)} & 71.4 & 8 & 72 & 62.4 & 11 & 116 \\
         &  & 2-2-2-2 & & 75.3 & 9 & 74 & 58.6 & 12 & 123 \\
        \midrule
        Conv-SE-NeXt (Ours) & 4 & 1-1-1-1 & FRNet Fusion~\cite{xu2025frnet} & 76.3 & 9 & 78 & 65.0 & 12 & 118 \\
        \bottomrule
        \multicolumn{10}{c}{\textbf{Our Baseline Model}} \\
        \toprule
        Conv-SE-NeXt (Ours) & 4 & 1-1-1-1 & Multi-scale (ours) & \textbf{77.1} & \textbf{7} & \textbf{61} & \textbf{65.1} & \textbf{10} & \textbf{107} \\
        \bottomrule
    \end{tabular}}
\end{table*}
Ultimately, our baseline model offers the best compromise between accuracy and efficiency.

\section{CONCLUSIONS}
\label{sec:conclusion}
In this work, we present HARP-NeXt, a high-speed and accurate range-point fusion network for real-time LiDAR semantic segmentation. By fusing 2D pixels and 3D points across multiple scales in a residual-attentive manner, HARP-NeXt effectively captures fine-grained details and broader contextual information, enhancing the robustness of predictions. Its core feature extraction module, Conv-SE-NeXt, integrates depth-wise separable convolutions with a squeeze-and-excitation mechanism to capture key patterns and reduce computations. Experiments on nuScenes and SemanticKITTI prove that HARP-NeXt achieves the best speed and accuracy trade-off on both high-end and embedded platforms.

\addtolength{\textheight}{-12cm}   

\bibliographystyle{IEEEtran} 
\bibliography{iros_main}

\end{document}